\pdfoutput=1

\documentclass[11pt]{article}

\usepackage[final]{acl}

\usepackage{xcolor}
\usepackage{soul}

\usepackage{times}
\usepackage{latexsym}
\usepackage{enumerate}
\usepackage[T1]{fontenc}

\usepackage[utf8]{inputenc}

\usepackage{microtype}

\usepackage{inconsolata}
\usepackage{pifont}

\usepackage{graphicx}
\usepackage{xcolor,xspace}
\usepackage{url}
\usepackage{multirow}
\usepackage{enumitem}
\usepackage{booktabs} 
\usepackage{amssymb}
\usepackage{paralist}
\setdefaultleftmargin{0pt}{}{}{}{}{}
\usepackage[framemethod=TikZ]{mdframed}

\usepackage{amsmath}
\usepackage{makecell}

\usepackage{amsthm}
\newtheoremstyle{mydef}
  {} 
  {} 
  {\normalfont} 
  {} 
  {\bfseries} 
  { } 
  { } 
  {\thmname{#1}~\thmnumber{#2} \textbf{#3}} 

\theoremstyle{mydef}

\newtheorem{definition}{Definition}[section]

\newcommand{\ignore}[1]{}

\newcommand{\descf}{{\ensuremath{\sf{\mathcal F}}}\xspace}
\newcommand{\mr}{{\ensuremath{\mathcal{\sf MR}}}\xspace}
\newcommand{\mrsameclient}{{\ensuremath{\mathcal{\sf MR_{Intra}}}}\xspace}
\newcommand{\mrdiffclient}{{\ensuremath{\mathcal{\sf MR_{Inter}}}}\xspace}
\newcommand{\mrcl}{{\ensuremath{\mathcal{\sf MR_{TotalCL}}}}\xspace}
\newcommand{\mrfl}{{\ensuremath{\mathcal{\sf MR_{TotalFL}}}}\xspace}
\newcommand{\mrpairwise}{\ensuremath{\mathcal{\sf MR}_{j \rightarrow k}}\xspace}
\newcommand{\mrpairwisejj}{{\ensuremath{\mathcal{\sf MR}_{j \rightarrow j}}}\xspace}

\title{Exploring Cross-Client Memorization of Training Data \\ in Large Language Models for Federated Learning}

\author{
 \textbf{Tinnakit Udsa\textsuperscript{1}},
 \textbf{Can Udomcharoenchaikit\textsuperscript{1}},
 \textbf{Patomporn Payoungkhamdee\textsuperscript{1}},
 \\
 \textbf{Sarana Nutanong\textsuperscript{1}},
 \textbf{Norrathep Rattanavipanon\textsuperscript{2}}
\\
\\
 \textsuperscript{1}School of Information Science and Technology, VISTEC
 \\
 \textsuperscript{2}College of Computing, Prince of Songkla University
 \\
 \texttt{\{tinnakit.u\_s24, canu\_pro, patomporn.p\_s21, snutanon\}@vistec.ac.th}
 \\
 \texttt{norrathep.r@phuket.psu.ac.th}
}

\begin{document}
\maketitle
\begin{abstract}

Federated learning (FL) enables collaborative training without raw data sharing, but still risks training data memorization. 
Existing FL memorization detection techniques focus on one sample at a time, underestimating more subtle risks of cross-sample memorization. 
In contrast, recent work on centralized learning (CL) has introduced fine-grained methods to assess memorization across all samples in training data, but these assume centralized access to data and cannot be applied directly to FL. 
We bridge this gap by proposing a framework that quantifies both intra- and inter-client memorization in FL using fine-grained cross-sample memorization measurement across all clients. 
Based on this framework, we conduct two studies: (1) measuring subtle memorization across clients and (2) examining key factors that influence memorization, including decoding strategies, prefix length, and FL algorithms.
Our findings reveal that FL models do memorize client data, particularly intra-client data, more than inter-client data, with memorization influenced by training and inferencing factors.
Code for the framework is available at \url{https://github.com/tinnakitudsa/FL_memorization_framework.git}.

\end{abstract}

\section{Introduction}
\label{sec:intro}

Federated learning (FL) allows collaborative model training across multiple clients without sharing raw data, thereby preserving privacy in domains like healthcare. 
However, FL does not eliminate the risk of memorization, where large language models (LLMs) may inadvertently encode sensitive data.
Prior work employs techniques such as canary injection, verbatim and exact memorization, k-extractible metrics, and BLEU score~\citep{canary, carlini2022quantifying, verbatim2, exact_mem, k_extractible, ippolito-etal-2023-preventing, kiyomaru-etal-2024-comprehensive-analysis} to measure memorization of LLMs in centralized learning (CL). 
However, these techniques share a critical assumption: a memorized text (suffix) can only be triggered by a prompt (prefix) from the same sample. 
This assumption does not hold in FL, where memorization may occur across clients (and thus across samples).
%
%
Recent techniques~\citep{ExploreMemFT,DoLMPlagiarize} lifted this assumption by proposing measurement of cross-sample memorization in CL. 
%

In contrast, existing research on FL memorization of LLMs has focused primarily on canary injection -- embedding out-of-distribution phrases into training data to see if the model reproduces them at inference time \citep{LSTMmemFLinjection, FLProductionInjection}. 
While these techniques can detect verbatim in the same sample, they are poorly suited to capturing more realistic leakage across samples of actual in-distribution training data.
%
Thus, FL memorization is likely to underestimate the real amount of memorization of training data.

This work is centered around the question: \emph{How can we adapt cross-sample memorization assessment from CL to measure realistic memorization risks in FL, and what factors influence such leakage?}
To address this question, we propose a framework that measures cross-client memorization in FL. 
The crux of our framework lies in a new \emph{pairwise} technique that extends prior CL methods to estimate memorization between FL clients.

Based on this framework, we then formulate the following studies to answer the above question. 
First, we assess the extent to which the global model in an FL setup memorizes client training data in a cross-client fashion, capturing both intra- and inter-client memorization. 
%
Second, we empirically analyze the potential factors that may affect memorization, including decoding methods, prefix length, federated algorithm, model size, and the number of communication rounds.
Together, these studies offer a comprehensive view of memorization risks across all clients in FL and show how CL insights can be adapted to decentralized contexts.

%
\ignore{Our findings reveal that both types of memorization occur across all tasks, with the intra-client consistently higher, indicating that the global model retains more information from the same client that supplied the prompt.}
%
%

The main contributions of this study are outlined as follows.
\textbf{i) Problem Formulation:} We frame the challenge of detecting subtle memorization in FL by adapting cross-sample assessment methods from CL.
\textbf{ii) Framework:} We design a framework that measures intra- and inter-client memorization in FL across clients, and conduct studies to quantify its extent and influencing factors.
\textbf{iii) Key insights:} 
FL memorization occurs across clients, but its effect is less pronounced than within the same client. 
The memorization degree depends on the prefix length, the decoding strategy, and the federated algorithm. 
However, we did not observe a clear trend for model sizes and communication rounds.
Additionally, the lengths of training inputs and outputs influence the length of generated text, which can limit the extractability of memorization in short-output tasks (e.g., classification).
Finally, suffixes associated with certain clients appear to be more susceptible to memorization than those of others.


\section{Memorization in Centralized Learning}
\label{sec:memorization}

We consider a CL setting in which a server trains an LLM $M$ on a local dataset $D$ consisting of $N$ text samples, i.e., $D = \{d_i\}_{i=1}^N$, where
$d_i$ is split into a fixed-length prefix $p_i$ and suffix $s_i$ such that $d_i = p_i || s_i$. Let $P$ and $S$ denote the set of all prefixes and suffixes, i.e., $P = \{p_i\}_{i=1}^N$ and $S = \{s_i\}_{i=1}^N$, prior work~\cite{ExploreMemFT} defines memorization as:

\begin{definition}[In-distribution CL memorization]
\label{def:cl-mem}

    A prefix $p \in P$ is said to induce memorization of a model $M$ in its in-distribution training data if there exists suffix $s \in S$ such that $\descf(M(p), s) = \text{True}$ for some discriminative function \descf that determines the similarity between two texts.
\end{definition}

Then, \mr, the memorization ratio of $M$ on its training data, can be computed as the fraction of prefixes in $P$ satisfying Definition~\ref{def:cl-mem}.

In addition to cross-sample memorization, \citet{ExploreMemFT} adapts fine-grained memorization measurement that goes beyond exact matches. 
They instantiated \descf using the PAN2014 plagiarism detector~\citep{SanchezPerez2014Winning, SanchezPerez2015Adaptive} 
measuring text similarity at three levels of granularity: (i) \textbf{verbatim}, (ii) \textbf{paraphrase} and (iii) \textbf{idea-level}.
%

\section{Proposed Study}
\label{sec:study}

\subsection{Memorization in Federated Learning}
\label{ssec:mem_FL}

In FL, we consider $L$ clients, where each client $C_i$ holds a private local dataset $D_i$.
Similar to CL, each $D_i$ can be divided into prefixes $P_i$ and suffixes $S_i$.
The aim in FL is to collaboratively train a shared global model $M$, without directly sharing any $D_i$.
FL background is further provided in Appendix~\ref{sec:appendix_fl}.

To define in-distribution memorization in FL, we propose to generalize Definition~\ref{def:cl-mem} as follows:

\begin{definition}[In-distribution FL memorization]\label{def:fl-mem}
    A prefix $p_j \in P_j$ from client $C_j$ is said to induce memorization of $M$ on the in-distribution training data of client $C_k$ if there exists suffix $s_{k} \in S_k$ such that $\descf(M(p_j),s_k)=\text{True}$ for a discriminator \descf.

\end{definition}

\begin{figure*}[htbp]
\centering
\includegraphics[scale=0.43]{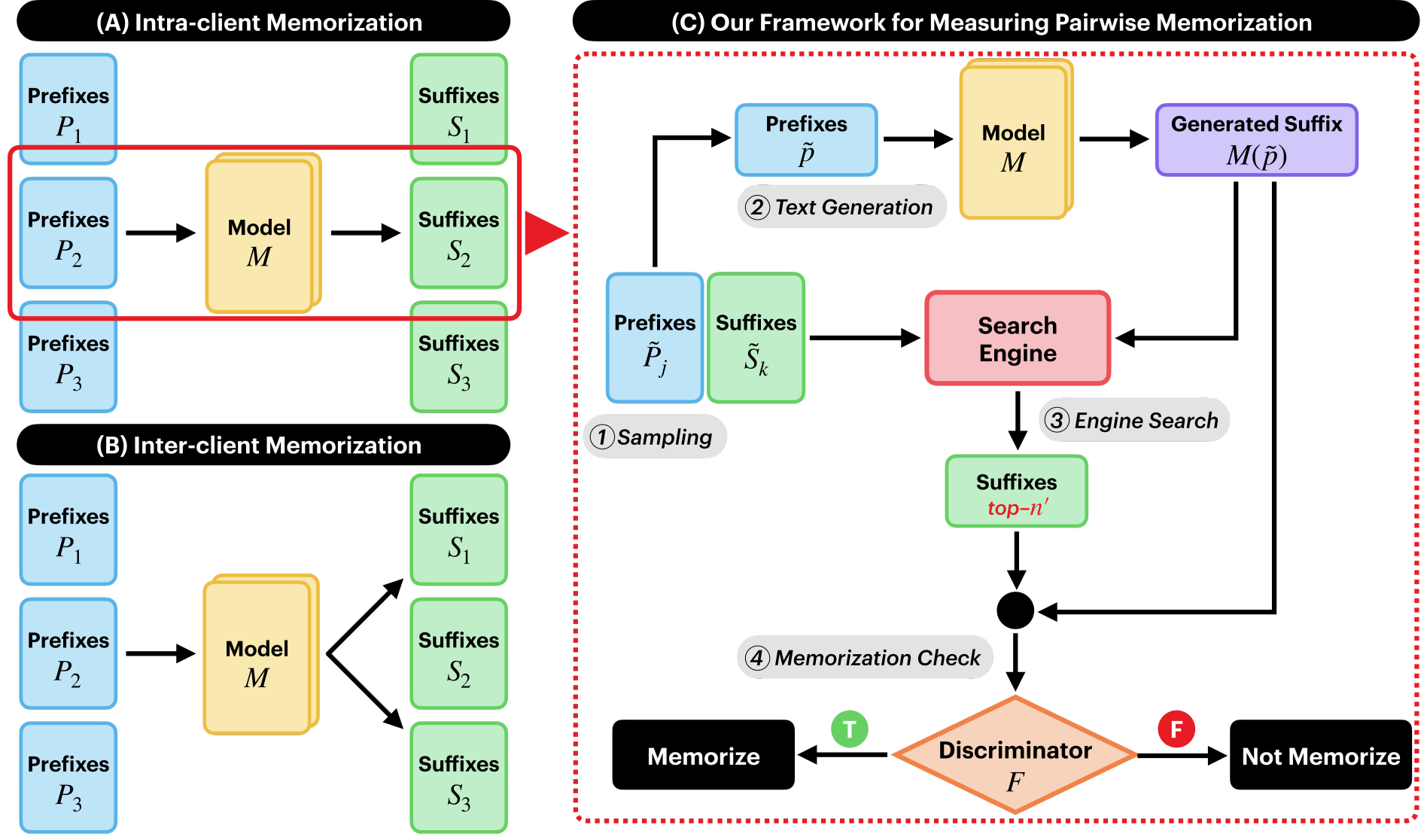}
\caption{Harm-exposed memorization (A) uses same-client prefixes/suffixes as input to our framework (C) (adapted from~\citet{DoLMPlagiarize, ExploreMemFT}) while harmful memorization (B) uses those from different clients.}
\vspace{-2ex}
\label{fig:fl-mem}
\end{figure*}

We categorize FL memorization into two types.
The first, \emph{inter-client memorization}, occurs when $C_j \neq C_k$ (Figure~\ref{fig:fl-mem}B).
This type is considered {\it harmful} since $C_j$ can use its own prefix $p_j$ to extract a suffix $s_k$ belonging to another client $C_k$, leading to direct privacy leakage in FL.
The second type, \emph{intra-client memorization}, occurs locally on a client’s own data, i.e., $C_j = C_k$ in Figure~\ref{fig:fl-mem}A. 
While this does not directly violate inter-client privacy, it may still pose a risk if the client's prefixes are known to others (e.g., $P_j$ is public).
As such, we classify it as {\it harm-exposed} rather than harmful.

\noindent\textbf{Memorization Metrics.} 
FL memorization, unlike CL, occurs across clients.
We define a \emph{pairwise memorization ratio} (\mrpairwise) from client $C_j$ to $C_k$ as a fraction of prefixes from $C_j$ causing the model $M$ to memorize a suffix belonging to $C_k$:

{\small
\begin{equation}\label{eq:client-mr}
    \mrpairwise = \frac{|P_{j,k}|}{|P_j|}
\end{equation}
}

where $P_{j,k}$ denotes a subset in $P_j$ that induce memorization on $C_k$ w.r.t. Definition~\ref{def:fl-mem}.

Using \mrpairwise, we define \mrsameclient and \mrdiffclient as the memorization ratios for intra-client and inter-client memorizations, respectively, averaged across all client pairs. 
Intuitively, they reflect how often a prefix from one client causes the model to memorize a private suffix of either the same (\mrsameclient) or a different (\mrdiffclient) client. 
To compare CL vs FL memorization, we introduce \mrcl and \mrfl, which capture the \emph{total} number (in percentage) of memorization-inducing prefixes from all clients in CL and FL, respectively.
Appendix~\ref{sec:appendix_notion} provides formal definitions of these metrics. 

\noindent\textbf{Estimating FL Memorization.} 
Precise measurement of memorization in LLMs is a challenging task~\cite{carlini2022quantifying}. 
To tackle this, we extend the estimation method proposed by~\citet{ExploreMemFT, DoLMPlagiarize} to compute \mrpairwise.
Given prefixes $P_j$ from client $C_j$ and suffixes $S_k$ from $C_k$, our framework calculates \mrpairwise by estimating $P_j$ fraction that induces memorization on $S_k$.
Figure~\ref{fig:fl-mem} shows the steps in our framework.

In~\ding{192}, 
we sample $\tilde{P_j} \subset P_j$ and $\tilde{S_k} \subset S_k$ of size $n \ll \text{min}(|P_j|,|S_k|)$, where 
$n=4K$ in our experiments.
We use each prefix $\tilde{p} \in \tilde{P_j}$ as a prompt to $M$, generating a text $M(\tilde{p})$ in~\ding{193}. 
In~\ding{194}, we index $\tilde{S_k}$ using
Elasticsearch,
and query it with each $M(\tilde{p})$ to retrieve top-$n'$ suffixes from $\tilde{S_k}$ that are most similar to $M(\tilde{p})$; following \citet{ExploreMemFT}, $n'=10$.
In~\ding{195}, we compare $M(\tilde{p})$ with the top-$n'$ suffixes using the PAN2014 plagiarism detector, which acts as the discriminative function \descf.
Finally, in~\ding{196}, if the detector returns \textit{True} (from any of the verbatim, paraphrase or idea-level similarity types) for any top-$n'$ suffixes, we consider $\tilde{p}$ to induce memorization w.r.t. Definition~\ref{def:fl-mem} and estimate \mrpairwise as the fraction of prefixes in $\tilde{P_j}$ triggering \textit{True}.
\ignore{
PAN2014 supports 3 categories of plagiarism: (i) \textbf{verbatim} that returns \textit{True} when two text inputs $t_1$ and $t_2$ share an exact substring exceeding a specific length threshold; (ii) \textbf{paraphrase} that detects whether $t_1$ contains a paraphrased text of $t_2$; and (iii) \textbf{idea} that identifies whether $t_1$ conveys the main idea of $t_2$ in a more concise form.
}
Based on \mrpairwise, aforementioned memorization metrics can be derived (see Appendix~\ref{sec:appendix_notion}).
As a result, these metrics let us explore the following RQs.

\vspace{-1ex}
\subsection{Research Questions (RQs)}
\label{ssec:rq1}

\begin{center}
  \begin{mdframed}[linewidth=1pt, innerleftmargin=4pt,innerrightmargin=4pt]
    \centering
    \emph{RQ1: Do FL models memorize training data?}
  \end{mdframed}
  \vspace{-2ex}
\end{center}

In this RQ, we aim to empirically evaluate \mrsameclient and \mrdiffclient for LLMs trained under FL settings to better understand the associated privacy risks.
As byproducts, we also examine: (1) whether FL leads to higher/lower memorization compared to CL, i.e., whether $\mrcl > \mrfl$ and (2) how model accuracy influences memorization.

\begin{center}
  \begin{mdframed}[linewidth=1pt, innerleftmargin=4pt,innerrightmargin=4pt]
    \centering
    \emph{RQ2: Which factors impact FL memorization?}
  \end{mdframed}
  \vspace{-2ex}
\end{center}

As shown in prior studies~\citep{ExploreMemFT, DoLMPlagiarize, kiyomaru-etal-2024-comprehensive-analysis, carlini2022quantifying}, the extent of CL memorization can be attributed to several factors.
Shifting from CL to FL, we aim to investigate whether similar patterns hold.
To this end, we perform ablation studies to examine the impact of both existing factors 
and FL-specific factors on FL memorization.

\subsection{Experimental Setup}
\textbf{Datasets.}
We evaluate our approach on four tasks: summarization, dialog, question answering (QA), and classification.
For each task, we consider 3 FL clients, where each client contains a distinct dataset originating from a domain with potential risks of sensitive information leakage. Detailed descriptions are provided in Appendix~\ref{sec:appendix_dataset}.

\noindent \textbf{Models.} 
Our main experiments are conducted using Qwen2.5-3B \cite{qwen2025qwen25technicalreport}.
In addition, we show the results for Llama3.2 \cite{grattafiori2024llama3herdmodels} and GPT2 \cite{gpt2}, along with further results of model size variations, in Appendix~\ref{sec:detailed_result}.

\noindent\textbf{Default Setting.} Unless stated otherwise, we report the results using the following setup: Qwen2.5-3B with prefix length of 30, top-k decoding method, and 3 FL rounds trained under FedAvg~\citep{fedavg}. Detailed configurations are shown in Appendix~\ref{sec:appendix_ablation_setting}.

\section{Results}
\label{sec:results}


\subsection{Memorization in Federated Learning}
\label{ssec:main_results}

As shown in Table~\ref{tab:small_main_table},
the models do memorize training data even in FL settings. In particular, \mrsameclient is consistently higher than \mrdiffclient across all evaluated tasks.
This answers \emph{RQ1} that FL memorization tends to occur more within the same client than across clients.
Interestingly, the classification task shows no memorization.
Further analysis of this behavior, including correlations between input and output lengths, is provided in Section~\ref{sec:analysis}.

\begin{table}[ht]
\small
\vspace{-2ex}
\caption{Intra- vs inter-client memorization in FL}
\label{tab:small_main_table}
    \vspace{-3ex}
    \begin{center}
        \begin{tabular}{lcc} 
        \toprule

            Tasks &             \mrsameclient(\%) &     \mrdiffclient(\%) \\ 
            \hline
            \hline
            Summarization &     0.342 &             0.046 \\
            Dialog &           1.533 &             1.446  \\
            QA &                1.450 &             0.813 \\
            Classification &    0.000 &             0.000 \\
        \bottomrule
        \end{tabular}
    \end{center}
    \vspace{-2ex}
\end{table}
%
Table~\ref{tab:small_trade_off} shows no clear trend in memorization when shifting from CL to FL. 
%
This finding contrasts with \citet{LSTMmemFLinjection}, which reported reduced memorization in FL. 
We expect this difference arises from the setup of memorization measurements: their work evaluates out-of-distribution memorization (via canary injection), whereas ours focuses on the in-distribution one.
%
Additionally, we found no correlation between memorization and performance in either setting.



\begin{table}[h]
\small
    \vspace{-2ex}
\caption{Trade-off between CL and FedAvg}
\label{tab:small_trade_off}
\vspace{-3ex}
    \begin{center}
        \begin{tabular}{lcc|c} 
        \toprule
                       Task &  Models  & Performance & \makecell{Memori- \\ zation(\%)}  \\
                       \hline
                       \hline
 \multirow{2}{*}{\makecell{Summa-\\zation}}   & \mrcl & 28.46 & 0.558 \\
                        & \mrfl  & 29.88  & 0.433 \\

                       \hline
\multirow{2}{*}{Dialog}   & \mrcl  & 19.40 & 3.417 \\
                        & \mrfl  & 18.11 & 3.992 \\

                       \hline
 \multirow{2}{*}{QA}   & \mrcl  & 26.66 & 2.150 \\
                        & \mrfl  & 28.60 & 2.917 \\
                    
                       \hline
\multirow{2}{*}{\makecell{Classi- \\ fication}}   & \mrcl & 76.30  & 0.000 \\
                        & \mrfl & 51.22  &  0.000 \\

        \bottomrule
        \end{tabular}
    \end{center}
    \vspace{-2ex}
\end{table}

Llama3.2-3B shows similar results to Qwen2.5-3B; see Appendix~\ref{sec:detail_per_cat_result} and~\ref{sec:detail_trade_off_result} for detailed results.
\subsection{Memorization Factors}

Next, to answer \emph{RQ2}, we investigate the potential factors that influence \mrsameclient and \mrdiffclient.

\begin{table}[ht]
    \vspace{-1ex}
\small
\caption{Memorization with various decoding methods}
\label{tab:small_decode_table}
    \vspace{-3ex}
    \begin{center}
        \begin{tabular}{lccc} 
        \toprule
    Tasks   & Decoding &  \mrsameclient(\%) & \mrdiffclient(\%) \\
                        \hline
                        \hline
    \multirow{3}{*}{\makecell{Summa-\\zation}  }   
                                        & temperature   & 0.475 & 0.067 \\
                                        & top-k         & 0.342 & 0.046 \\
                                        & top-p         & 0.525 & 0.050 \\
                                        \hline
    \multirow{3}{*}{Dialog}          
                                        & temperature   & 1.267 & 1.442 \\
                                        & top-k         & 1.533 & 1.446 \\
                                        & top-p         & 3.792 & 2.996 \\
                                        \hline
    \multirow{3}{*}{QA}              
                                        & temperature   & 1.283 & 0.750 \\
                                        & top-k         & 1.450 & 0.813 \\
                                        & top-p         & 2.567 & 1.438 \\

        \bottomrule
        \end{tabular}
    \end{center}
    \vspace{-4ex}
\end{table}
\paragraph{Decoding Method.}
\label{ssec:decode}

Table~\ref{tab:small_decode_table} indicates
memorization tends to increase when applying top-k or top-p decoding.
The results align with prior work~\citep{ExploreMemFT, DoLMPlagiarize}, where sophisticated decoding strategies amplify memorization effects.

\noindent\textbf{Prefix Length.}
\label{ssec:prefix_length}
As shown in Table~\ref{tab:small_prefix_table}, increasing prefix length lowers \mrdiffclient and \mrsameclient in most cases, indicating that shorter prefixes can induce greater memorization in FL.

\begin{table}[ht]
\small
\vspace{-2ex}

\caption{Memorization with various prefix lengths}
\label{tab:small_prefix_table}
    \vspace{-3ex}
    \begin{center}
        \begin{tabular}{lccc} 
        \toprule

    Tasks   & \makecell{Prefix Length} &  \mrsameclient(\%) & \mrdiffclient(\%) \\
                        \hline
                        \hline
    \multirow{3}{*}{\makecell{Summa-\\zation}}     
                                        & 10            & 0.508 & 0.188 \\
                                        & 30            & 0.342 & 0.046 \\
                                        & 50            & 0.425 & 0.038 \\
                                        & 100           & 0.208 & 0.004 \\
                                        \hline
    \multirow{3}{*}{Dialog}          
                                        & 10            & 2.108 & 1.992 \\
                                        & 30            & 1.533 & 1.446 \\
                                        & 50            & 1.575 & 1.242 \\
                                        & 100           & 1.408 & 1.150 \\
                                        \hline
    \multirow{3}{*}{QA}              
                                        & 10            & 1.525 & 1.383 \\
                                        & 30            & 1.450 & 0.813 \\
                                        & 50            & 1.242 & 0.550 \\
                                        & 100           & 1.125 & 0.429 \\

        \bottomrule
        \end{tabular}
    \end{center}
    \vspace{-5ex}
\end{table}

\paragraph{Federated Algorithm.}
\label{ssec:fl_algo}
Table~\ref{tab:small_fed_algo_table} indicates that FedProx leads to higher memorization rates (\mrdiffclient and \mrsameclient).
Notably, it exhibits memorization in the classification task, which is absent in FedAvg. 
This indicates that memorization varies across algorithms and should be evaluated independently.

\begin{table}[ht]
\small
\vspace{-1ex}

\caption{Memorization across federated algorithms}
\label{tab:small_fed_algo_table}
    \vspace{-3ex}
    \begin{center}

        \begin{tabular}{lccc} 
        \toprule
    Tasks   & \makecell{Federated  \\ Algorithm} &  \mrsameclient(\%) & \mrdiffclient(\%) \\
                        \hline
                        \hline
    \multirow{2}{*}{\makecell{Summa-\\zation}}     
                                        & FedAvg            & 0.342 & 0.046 \\
                                        & FedProx           & 0.942 & 0.138 \\
                                        \hline
    \multirow{2}{*}{Dialog}          
                                        & FedAvg            & 1.533 & 1.446 \\
                                        & FedProx           & 1.892 & 1.879 \\
                                        \hline
    \multirow{2}{*}{QA}              
                                        & FedAvg            & 1.450 & 0.813 \\
                                        & FedProx           & 3.675 & 2.146 \\
                                        \hline
    \multirow{2}{*}{\makecell{Classi-\\fication}}              
                                        & FedAvg            & 0.000 & 0.000 \\
                                        & FedProx           & 0.011 & 0.000 \\

        \bottomrule
        \end{tabular}
    \end{center}
    \vspace{-3ex}
\end{table}

Besides the aforementioned factors, our experiments also consider model size and the communication rounds.
The results show no strong association with \mrdiffclient and \mrsameclient; detailed results are provided in Appendix~\ref{sec:full_size} and~\ref{sec:full_comm_round}.










\section{Observations on Memorization Patterns}
\label{sec:analysis}

Here, we conduct two analyses to gain deeper understanding of memorization in FL.

\subsection{Correlations Between Input Length, Output Length, and Generated Suffix Length}
\label{ssec:gen_length}

\begin{table}[h]
\vspace{-2ex}
\small
\caption{Token length of input, output, and generated text. (Median)}
\label{tab:generation}
\tabcolsep=0.11cm
\vspace{-2ex}
    \begin{center}
        \begin{tabular}{lccc|c} 
        \toprule
                       Model & Task &  Input & Output &  \makecell{Generated \\ Text}   \\
                       \hline
                       \hline
\multirow{4}{*}{Qwen2.5-3B} & { \makecell{Summarization}}   & 184  & 15 & 10 \\
                        & {Dialog}   & 86  & 108 & 59 \\
                        & {QA}   &  325 & 47 & 78 \\
                        & {Classification}   & 38  & 2 & 1 \\
                       \hline
\multirow{4}{*}{Llama3.2-3B} & { \makecell{Summarization}}   & 182  & 15 & 6 \\
                        & {Dialog}   &  84 & 107 & 66 \\
                        & {QA}   & 313  & 47 & 26 \\
                        & {Classification}   & 37  & 2 & 2 \\
                       \hline
        \bottomrule
        \end{tabular}
    \end{center}
    \vspace{-2ex}
\end{table}

First, we analyze the correlations between training input length, training output length, and generated suffix length (from our memorization framework). 
Table~\ref{tab:generation} shows that the generated suffix length is strongly affected by the input and output lengths
Specifically, in summarization and classification tasks, the median of generated suffix lengths is close to the output length. 
This phenomenon leads the classification task to generate a smaller token than other tasks to the point that its generated suffix length is smaller than the memorization threshold set in PAN2014, leading to 0.000\% memorization in most of our experiments.

\subsection{Effects of Suffixes}
\label{ssec:suffixes}


\begin{table}[h]
\vspace{-1ex}
\small
\caption{Effect of Suffix Index(\%)}
\label{tab:suffixes_table}
\vspace{-2ex}
    \begin{center}
        \begin{tabular}{ccc}
        \toprule
             Prefix & Suffix & \mrpairwise \\
        \hline\hline
             Group1 & Group1 & 1.450 \\
             Group1 & Group2 & 1.525 \\
             Group1 & Group3 & 1.500 \\
             Group2 & Group1 & 1.150 \\
             Group2 & Group2 & 1.200 \\
             Group2 & Group3 & 1.225 \\
             Group3 & Group1 & 1.725 \\
             Group3 & Group2 & 1.550 \\
             Group3 & Group3 & 1.950 \\
        \bottomrule
        \end{tabular}
    \end{center}
\vspace{-1ex}
\end{table}

It is worth noting from Table~\ref{tab:suffixes_table} that memorization is more likely when the prefix and suffix are from the same client. Surprisingly, suffixes from Group3 of Dialog task (see Appendix~\ref{ssec:dialog_data}) are memorized more than those from other clients. This suggests that the characteristic of a dataset to which suffixes belong plays an important role in their chance of being memorized. Future research is needed to better understand which specific dataset properties drive this effect and how they interact with FL setting.

\section{Conclusion}
\label{sec:conclusion}

We presented a framework for evaluating fine-grained cross-sample and cross-client memorization in FL, adapting 
%
techniques from CL to measure both intra- and inter-client leakage. 
Empirical results show that FL models tend to memorize training data, with the intra-client memorization consistently higher than the inter-client memorization in all cases. 
Among all factors examined, our findings indicate that 
FL memorization rates depend on prefix length, decoding strategy, and federated algorithm.
Also, we find no clear evidence that memorization is reduced in FL compared to CL, emphasizing the need to quantify privacy leakage even when privacy-sensitive applications are trained under FL.


\section*{Limitations}
Despite its contributions, this study has limitations. Firstly, the PAN2014 plagiarism detector can produce misleading results when models generate incoherent output (see Appendix~\ref{ssec:gibberish}). 
Note that, for the results reported in our paper, we use models that can generate coherent outputs. 
Secondly, the PAN2014 plagiarism detector, can only measures fine-grained memorization in English texts.
Lastly, while this study provides empirical results, theoretical investigation into why intra-client memorization exceeds inter-client memorization is an interesting research direction. 
%
%

\section*{Acknowledgments}
Computing resources were supported part by the ThaiLLM collaborations, funded by the Digital Economy and Society Development Fund of the Ministry of Digital Economy and Society, Thailand; the Fundamental Fund (FF 87245) from Thailand Science Research and Innovation (TSRI), 2025; and the WangchanX project through donations from SCB, SCBX, and PTT.
%
%
Appreciation is extended to members of the Natural Language Processing and Representation Learning Lab at VISTEC for their technical and moral support, especially Wuttikorn Ponwitayarat, Pume Tuchinda, and Natnasa Lertmahakul.
The first author thanks the co-authors and co-advisors for their guidance and insightful feedback throughout this work,
friends and family for their encouragement and understanding,
and himself for the perseverance and dedication that made this work possible.

\bibliography{anthology,References}

\clearpage

\section*{Appendix}
\appendix
\section{Federated Learning}
\label{sec:appendix_fl}

The goal of FL is to collaboratively train a model from multiple sources without directly sharing their data. The widely adopted FL framework is FedAvg~\citep{fedavg}. 
In FedAvg, each client trains a local model on its own dataset. 
Afterward, it sends the locally trained model to a central server where all local models are aggregated via averaging.
The averaged model is sent back to local clients to continue training, repeating this process iteratively. 
Ultimately, the globally averaged model is used for all clients.

Another FL algorithm used in this work is FedProx~\citep{fedprox}.
It addresses the heterogeneity issue of FedAVG (e.g., non-IID data) by modifying the local objective of each client to include a proximal (regularization) term that penalizes deviations from the current global model.

\section{Datasets}
\label{sec:appendix_dataset}

To investigate the impact of fine-tuning in FL, we conduct separate experiments with publicly available datasets on summarization, dialog, question-answering, and classification tasks. 
Each dataset is partitioned among clients in a non-IID fashion. Task-specific details are provided below.


\subsection{Summarization}
\label{ssec:summmarization_data}
Following \citet{DoLMPlagiarize}, we use the ArxivAbstract~\citep{arxiv_archive} dataset, as too much memorization in academic writing can be seen as a form of plagiarism. The dataset contains abstracts collected from arXiv.org. We focus on three subject areas: Astrophysics, Condensed Matter, and Mathematics. For each area, we sample 30,000 examples, with 27,000 used for training and 3,000 for testing. The subcategories are as follows:
\begin{compactenum}[i)]
\item \textbf{Astrophysics:} High Energy Astrophysical Phenomena, Instrumentation and Methods for Astrophysics, Earth and Planetary Astrophysics, and Astrophysics of Galaxies.
\item \textbf{Condensed Matter:} Superconductivity, Soft Condensed Matter, Disordered Systems and Neural Networks, Quantum Gases, and Other Condensed Matter.
\item \textbf{Mathematics:} K-Theory and Homology, Statistics Theory, Differential Geometry, Mathematical Physics.
\end{compactenum}

This dataset is publicly available at \url{https://github.com/staeiou/arxiv_archive}.


\subsection{Dialog}
\label{ssec:dialog_data}
We use the data from \citet{med_dialog}, consisting of 100,000 real patient-doctor dialogues sourced from HealthCareMagic.com.
We divide the dataset into 3 disjoint groups.
For each group, we sample 30,000 instances with 27K/3K for the train-test split and assign them to each FL client.
The groups are defined as follows:

\begin{compactenum}[i)]
\item \textbf{Group 1:} Child Health, Lung and Chest disorders, Mental Health, Cancer, Birth Control, Bones, Muscles and Joints, Natural and Home Remedies, and Accident and Emergency.
\item \textbf{Group 2:} Brain and Spine, Infections, Abdominal Pain, Liver and Gall Bladder, Kidney Conditions, Infertility Problem, Asthma and Allergy, Diabetes, and Lupus.
\item \textbf{Group 3:} Men's Health, Hypertension and Heart Disease, Dental Health, Lump, Pregnancy, Pain Management, Medicines and Side Effects, and Alternative Medicine.
\end{compactenum}

The dataset can be found in \url{https://github.com/tangg555/meddialog}.

   
\subsection{Question-Answering}
\label{ssec:qa_data}
We choose PubMedQA~\citep{jin2019pubmedqa}, the medical domain dataset, as the risk of sensitive information leakage is particularly critical in this field.
The dataset contains over 273,000 biomedical question-answering instances, including 1,000 expert-labeled, 61,200 unlabeled, and 211,300 artificially generated samples, all derived from PubMed abstracts. 
The task involves answering questions with one of three labels: yes, no, or maybe. 
Each instance includes both a short answer (the label) and a long-form explanation. 
In our experiments, we use the long answers as the ground truth labels. The dataset is available at: \url{https://huggingface.co/datasets/bigbio/pubmed_qa}.

We divide the dataset into 3 groups based on Medical Subject Headings. Each clients contain 30,000 sampled instances, with 27,000 used for training and 3,000 for testing. 
The group details are as follows:

\begin{compactenum}[i)]
\item \textbf{Group 1:} having no Humans tag and having Animals tag.
\item \textbf{Group 2:} having no Animals tag, having Humans tag, and having Middle Aged tag.
\item \textbf{Group 3:} having no Animals tag, having Humans tag, and having no Middle Aged tag.
\end{compactenum}

\begin{table*}[h]
\caption{The chat templates for each task employed in fine-tuning the LLMs and in subsequent evaluation.}
\label{tab:training_examples_table}
\tabcolsep=0.11cm
    \begin{center}
        \begin{tabular}{p{15cm}} 
\toprule
Summarization  \\
\hline
\textbf{User:} Please summarize the following abstract into a title. \\
\{Abstract\}

\textbf{Assistant:} \{Title\} \\

\toprule
Dialog  \\
\hline
\textbf{User:} If you are a doctor, please answer the medical questions based on the patient’s description.  \\
\{Patient\}

\textbf{Assistant:} \{Doctor\}
\\
\toprule
Question Answering (QA)  \\
\hline
\textbf{User:} \{Question\} \\
\{Context\}

\textbf{Assistant:} \{Answer\}
\\
\toprule
Classification  \\
\hline
\textbf{User:} Please classify the following passage into one of the following categories: BACKGROUND, OBJECTIVE, METHODS, RESULTS, or CONCLUSIONS. \\
\{Passage\}

\textbf{Assistant:} \{Class\}
\\
\bottomrule
        \end{tabular} 
    \end{center}
\end{table*}

\subsection{Classification}
\label{ssec:classification_data}
We use the PubMed 200k RCT~\citep{dernoncourt-lee-2017-pubmed} dataset, derived from PubMed, for sequential sentence classification. It contains around 200,000 abstracts from randomized controlled trials.
It consists of 5 components: background, objective, methods, results, and conclusion.
This dataset is publicly available at \url{https://github.com/Franck-Dernoncourt/pubmed-rct}.
Following standard practice of data partition for classification task in FL~\citep{fedbabu, fedala}, we partition the dataset among FL clients using a Dirichlet distribution with $\alpha =5.0$, where each client receives 30K instances with 27K/3K train-test split.


\subsection{Sample Format}
\label{ssec:format}

\label{sssec:training_format}

The pre-processed data used for model training are formatted according to Table~\ref{tab:training_examples_table}.
The formatted data are used to train the model with its corresponding template applied.


\section{Training Costs and Hyperparameters}
\label{sec:appendix_ablation_setting}

\subsection{Training Cost}
We used an NVIDIA A100 GPU with 4,000 GPU hours to train Qwen2.5-0.5B, Qwen2.5-1.5B, Qwen2.5-3B, Llama-3.2-1B, and Llama-3.2-3B. It takes the total of 1,600 GPU hours and 1,600 CPU hours to apply our memorization framework across all experiments.

\subsection{Hyperparameters}
We trained models using LLaMA Factory library\footnote{\url{https://github.com/hiyouga/LLaMA-Factory}} with a 2e-4 learning rate, bfloat16 and batch sizes of 64.

\subsection{Memorization Factors}

\paragraph{Decode Method:} Following \citet{DoLMPlagiarize}, we empirically study multiple decoding methods: top-k, top-p, and temperature. For each decoding, we start with the default Huggingface trainer-generated parameters and update them with the following: \texttt{k=40} for top-k, \texttt{p=0.8} for top-p, and \texttt{temperature=1.0} for temperature decoding.

\paragraph{Prefix Length:} Our experiments are conducted with different prefix lengths: 10, 30, 50, and 100 tokens, following~\citet{ExploreMemFT}.

\paragraph{Federated Algorithm:} We use FedAvg and FedProx as federated algorithm to train and measure memorization.

\paragraph{Communication Rounds:} We conduct experiments using 1, 3, and 5 federated communication rounds, which effectively correspond to the number of training epochs in the FL process.

\paragraph{Model Size:} We select Qwen2.5-0.5B, Qwen2.5-1.5B, Qwen2.5-3B, Llama-3.2-1B, adn Llama-3.2-3B, which have 494M, 1.54B, 3.09B, 1.24B, and 3.21B parameters, respectively, to measure memorization.


\section{Formal Definitions}
\label{sec:appendix_notion}

\ignore{
\subsection{Pairwise Memorization Ratio}\label{apdx:mem-def-fl}
Let $P_{j,k}$ denote the set of all prefixes from client $C_j$ that induce memorization on client $C_k$'s training data according to Definition~\ref{def:fl-mem}.
We define a pairwise memorization ratio from $C_j$ to $C_k$, \mrpairwise, as a fraction of prefixes from $C_j$ causing the model $M$ to memorize a suffix belonging to $C_k$:

{\small
\begin{equation}\label{eq:client-mr}
    \mrpairwise = \frac{|P_{j,k}|}{|P_j|}
\end{equation}
}
}

\subsection{Harm-exposed and Harmful Memorization Ratios}\label{apdx:inter-intra-mem-def}
Based on the notion of \emph{intra-client (harm-exposed)} memorization and Equation~\ref{eq:client-mr}, we can define \mrsameclient, the overall harm-exposed memorization ratio, as a weighted sum on all $L$ clients:

{\small
\begin{equation}\label{eq:benign-mr}
    \mrsameclient = \sum_{j=1}^{L} w_j \cdot \mrpairwisejj
\end{equation}
}

where $w_j$ represents the weight of client $C_j$;
in our work, it is calculated as the ratio of $C_j$'s training dataset to the total training data among all $L$ clients, i.e., $w_j = |D_j| / \sum_{i=1}^{L} |D_i|$.

Then, we can compute the average harmful memorization ratio incurred by a specific client $C_j$ as:

{\small
\begin{equation}\label{eq:mal-mr}
    \mrdiffclient(j) = \frac{1}{L-1} \sum_{j \ne k} \mrpairwise
\end{equation}
}

and finally \mrdiffclient, the overall harmful memorization ratio, across all clients as a weighted sum:

{\small
\begin{equation}\label{eq:mal-mr}
    \mrdiffclient = \sum_{j=1}^{L} w_j \cdot \mrdiffclient(j)
\end{equation}
}

In our experiments, we approximate \mrsameclient and \mrdiffclient following the estimated \mrpairwise produced from the framework in Section~\ref{ssec:mem_FL}.

\subsection{Total Memorization Ratios in CL and FL}
To enable a fair comparison between CL and FL, we use a metric based on the \emph{total} number of prefixes that 
cause the model to memorize \emph{any} suffix.
In FL, this number corresponds to the union\footnote{Note that we take a \emph{union} rather than a summation to avoid double-counting memorization-inducing prefixes as a single prefix may cause the model to leak suffixes belonging to multiple clients, i.e., it is possible that $P_{j,k1} \cap P_{j,k2} \neq \varnothing$ for $k1 \neq k2$.} of all $P_{j,k}$-s. Consequently, we define the total memorization ratio in FL as:

{\small
\begin{equation}\label{eq:mrfl}
    \mrfl = \frac{|\bigcup_{j,k} P_{j,k}|}{|\bigcup_j P_j|} 
\end{equation}
}

The total memorization ratio in CL, \mrcl, is computed the same way as \mrfl via Equation~\ref{eq:mrfl}.
The key distinction, however, lies in how the model $M$ is trained:
in CL, $M$ is trained centrally on the combined datasets by a trusted third party, whereas in the FL setting, $M$ is trained in a distributed manner across clients.

\section{PAN2014 and Three Categories of Memorization}
\label{sec:measurement_detailed}

\subsection{PAN2014 Plagiarism Detector}
\label{ssec:pan2014}
PAN2014 plagiarism detector evaluates text similarity across three distinct categories: verbatim, paraphrase, and idea. This allows for fine-grained assessment of memorization that extends beyond exact matches captured by verbatim similarity. For this work, we use the improved version of PAN2014 proposed in~\citep{DoLMPlagiarize}. Detailed procedures for measuring cross-sample memorization using PAN2014 can be found in \citet{DoLMPlagiarize, ExploreMemFT}. 

\subsection{Memorization Category}
\label{ssec:mem_cat}
In this work, we use all 3 PAN2014 categories in our analysis. In PAN2014, paraphrasing is evaluated using RoBERTa and NER models, with predictions categorized as low-confidence when p < 0.5 and high-confidence when p > 0.5; as in~\citet{ExploreMemFT}, we report results for both categories in the breakdown provided in Appendix~\ref{sec:detailed_result}.

\subsection{Memorization Measurement}
\label{ssec:add_only_prefix}
To measure memorization, we use the abstract, patient, context, and passage texts (see Table~\ref{tab:training_examples_table}) as input (prefix+suffix) to our framework for the summarization, dialog, question-answering, and classification tasks, respectively.

\subsection{Hyperparameters for Memorization Measurement}
\label{ssec:hyper_pan}
Following \citet{ExploreMemFT}, we set the minimal match threshold to at least 50 characters for PAN2014 in all memorization types. 
We also filter out some odd cases, as described in Appendix~\ref{ssec:gibberish}, using a naive filter that excludes samples with at least ten repetitions of a three-word sequence. 

\begin{table*}[h]
\vspace{-2ex}
\small
\caption{Misclassify Memorization of Incoherent Output Generation}
\label{tab:gibberish_table}
\tabcolsep=0.11cm
\vspace{-2ex}
    \begin{center}
        \begin{tabular}{p{2.5cm} |p{6.5cm}| p{6.5cm}} 
        \toprule
                       Type &  Machine-Written Text  & Training Text  \\
                       \hline
                                Idea &   Algebraic Groups for Lie groups, Lie Groups and subalgebras, Lie algebras c to Lie algebras and Lie algebras such algebras, cohomologies, algebraic groups, Lie groups and Lie algebras without equations & Here we focus on contractions of Lie algebras and algebraic groups.    \\ \hline
                                   Idea &   Your thyroid is a butterfly-shaped gland at the base of your throat. You have two lobes, lobes, and three lobes. It has two lobes, two lobes, one lobe, and two lobes, one lobe, and two lobes, two lobes, and two lobes, both lobes at the base of your throat. It's involved in some very big jobs, like managing your heart rate, blood pressure, body temperature, and weight. Your thyroid has two lobes, lobes, lobes, lobes, lobes, lobes, lobes, lobes, lobes, lobes, and weight. & the thyroid gland is removed. A lobectomy is when one of the two lobes of your thyroid is removed.    \\

        \bottomrule
        \end{tabular} 
    \end{center}
    \vspace{-2ex}
\end{table*}

\subsection{Memorization Result of Incoherent Output Generation}
\label{ssec:gibberish}

When a model generates incoherent outputs (see Table~\ref{tab:gibberish_table}), the PAN2014 plagiarism detector misclassifies them as idea memorization. The reason for this phenomenon is that PAN2014 is designed to discriminate human-like texts, not incoherent texts.
As this assumption does not hold, it can leads to unpredictable results. This seems to be an apparent limitation for any PAN2014-based memorization methods.

\section{Memorized Examples}
\label{sec:examples}

\begin{table*}[h]
\small
\caption{Examples of memorization from FL}
\label{tab:examples_table}
\tabcolsep=0.11cm
    \begin{center}
        \small
        \begin{tabular}{p{2.5cm} |p{6.5cm}| p{6.5cm}} 
        \toprule
            Type &  Machine-Written Text  & Training Text  
            \\ \hline

            Verbatim &  ... I am concerned about my \hl{high blood pressure, high cholesterol, and triglycerides}. If they are high, what should be done?I can be managed by lifestyle changes, medications, and diet changes ...  &  ... I have been diagnosed with interstitial cystitis, fibromyalgia, mild psoriasis, \hl{high blood pressure, high cholesterol and triglycerides} and have frequent UTI's.  ...
            
            (Dialog, prefix: group1, suffix: group2)
            \\ \hline

            Verbatim &  in development \hl{of the mitogen-activated protein kinase (MAPK) pathway}. Reducing the expression of the Redd-1 protein may increase the levels of MAPK. In addition, increasing the expression of the Redd-1 protein may inhibit the progression of the MAPK pathway. The Redd-1 protein is known ...
            & ... We characterized the signaling properties of confirmed molecular alterations by ectopic expression of engineered cDNAs in 293H cells. Activation \hl{of the mitogen-activated protein kinase (MAPK) pathway} in cells by ectopic expression of PAPSS1-BRAF was abrogated by mitogen-activated protein  ...
            
            (QA, prefix: group1, suffix: group3)
            \\ \hline

            Verbatim 
            & comparison with a ground-based system \hl{for the medium-sized telescopes of the Cherenkov Telescope Array}
            & construction is scheduled to begin in fall at the Fred Lawrence Whipple Observatory in southern Arizona, USA. The Schwarzschild-Couder telescope is a candidate \hl{for the medium-sized telescopes of the Cherenkov Telescope Array}, which utilizes ... 
            
            (Summarization, prefix: astro, suffix: astro)
            \\ \hline
            Paraphrase &  do.  \hl{I have a strong medical history of cancer to my family.}  There are chances, but we shouldn't worry much as these are likely benign tumors  & of cancer, colon in 1996 and lung in 2005. \hl{I also have a strong history of cancer in my immediate family.} While at the dentist, I noticed a skin tag ... 
            
            (Dialog, prefix: group3, suffix: group1)
            \\ \hline

            Paraphrase 
            & PE) cellular functions including migration and apoptosis. We have developed a novel approach to study matrix effects on RPE, which provides the rationale for the treatment of many macular degenerative diseases through blocking active MMP. \hl{In this study, we investigated MMP effects on different cellular functions of RPE.}
             & Cinaciguat, the novel soluble guanylate cyclase activator, currently being in phase IIb clinical trial, has been shown to exert antiplatelet and anti-remodeling effects in animal models of vascular pathology. \hl{In this study we investigated the effects of cinaciguat on post-injury arterial stenosis.} Male Sprague-Dawley rats (n=100) underwent endothelial denudation ... 
             
             (QA, prefix: group3, suffix: group1)
            \\ \hline
            
            Paraphrase & (Target of Opportunity) missions to study the blazar source in depth. \hl{Fermi Large Area Telescope Target of Opportunity Missions}  & \hl{The data were taken with the Large Area Telescope on board the Fermi Gamma-ray Space Telescope.} An extended source is found at a position consistent with that of RCW 103, and its emission was only detected above 1 GeV ... 
            
            (Summarization, prefix: astro, suffix: astro)
            \\ \hline

            Idea &  \hl{am having very heavy throat pain and sometimes pain is felt in the chest , and sometimes it feels like something is pressing in the chest .i have cough and chest pain after eating rice .} i can feel the lump in the throat near tonsil and it is also painful ...
             &  clear from my throat. \hl{I also have a very sore throat and pain in my chest.} Is this a viral infection I just have to wait out or could there be something I can do about it? 
             
             (Dialog, prefix: group1, suffix: group2)
             \\ \hline

            Idea 
            &  HEK293 are the four cell lines analyzed. \hl{The expression of mRNA or protein can be measured by quantitative PCR. The expression level of rRNA (total rRNA) and total mRNA (mRNA) may represent the amount of TmRNA present in the cells. The rRNA expression level of mRNA in L929 cells could be measured by quantitative PCR, but not by quantitative RT (real time PCR) method.} ...
             &  knocked down by small interfering RNA (siRNA) in HeLa cells which were then cultured in conventional medium or serum starvation medium. The protein level of TXNDC5 was evaluated by Western blot analysis. \hl{The mRNA level of TXNDC5 was measured by quantitative real-time PCR.} Cell growth rate was determined by cell proliferation assay kit (MTS method). Cell cycle distribution and apoptosis were detected by flow cytometry. ...
             
             (QA, prefix: group1, suffix: group3)
             \\

        \bottomrule
        \end{tabular} 
    \end{center}
\end{table*}
Examples of memorization in the categories of verbatim, paraphrase, and idea are shown in Table~\ref{tab:examples_table}.

\section{Detailed Results}
\label{sec:detailed_result}

\subsection{Per-Category Results for \mrsameclient and \mrdiffclient}
\label{sec:detail_per_cat_result}
\begin{table*}[h]
\small
\caption{Per-Category Memorization in FL(\%)}
\label{tab:percat_mem_fl_table}
    \begin{center}
        \begin{tabular}{cccc|cccc}
        \toprule
              Model & Tasks & Type & Total & Verbatim & Idea & \makecell{Paraphrase\\ (p > 0.5)} & \makecell{Paraphrase\\(p < 0.5)}  \\
             \hline\hline
            \multirow{8}{*}{Qwen2.5-3B} 
             &  \multirow{2}{*}{\makecell{Summarization}} & \mrsameclient & 0.342 & 0.008 & 0.000 & 0.025 & 0.308 \\ 
             & & \mrdiffclient & 0.046 & 0.000 & 0.000 & 0.013 & 0.033 \\
             \cline{2-8}
             & \multirow{2}{*}{Dialog} & \mrsameclient & 1.533 & 0.000 & 0.067 & 0.258 & 1.217 \\
             & & \mrdiffclient & 1.446 & 0.000 & 0.042 & 0.292 & 1.125 \\
             \cline{2-8}
             & \multirow{2}{*}{QA} & \mrsameclient & 1.450 & 0.000 & 0.067 & 0.375 & 1.033 \\
             & & \mrdiffclient & 0.813 & 0.000 & 0.046 & 0.192 & 0.588 \\
             \cline{2-8}
             & \multirow{2}{*}{Classification} & \mrsameclient & 0.000 & 0.000 & 0.000 & 0.000 & 0.000 \\
             & & \mrdiffclient & 0.000 & 0.000 & 0.000 & 0.000 & 0.000 \\
             \hline             
            \multirow{8}{*}{Llama-3.2-3B} 
             &  \multirow{2}{*}{\makecell{Summarization}} & \mrsameclient & 0.700 & 0.008 & 0.008 & 0.125 & 0.558 \\ 
             & & \mrdiffclient & 0.067 & 0.000 & 0.000 & 0.013 & 0.054 \\
             \cline{2-8}
             & \multirow{2}{*}{Dialog} & \mrsameclient & 3.167 & 0.000 & 0.183 & 0.692 & 2.367 \\
             & & \mrdiffclient & 2.313 & 0.000 & 0.071 & 0.488 & 1.813 \\
             \cline{2-8}
             & \multirow{2}{*}{QA} & \mrsameclient & 2.183 & 0.083 & 0.050 & 0.583 & 1.500 \\
             & & \mrdiffclient & 1.375 & 0.008 & 0.046 & 0.367 & 0.983 \\
             \cline{2-8}
             & \multirow{2}{*}{Classification} & \mrsameclient & 0.000 & 0.000 & 0.000 & 0.000 & 0.000 \\
             & & \mrdiffclient & 0.000 & 0.000 & 0.000 & 0.000 & 0.000 \\
             \hline
            \multirow{8}{*}{GPT-2 XL} 
             &  \multirow{2}{*}{\makecell{Summarization}} & \mrsameclient & 0.567 & 0.008 & 0.067 & 0.075 & 0.417 \\ 
             & & \mrdiffclient & 0.075 & 0.000 & 0.017 & 0.008 & 0.050 \\
             \cline{2-8}             
             & \multirow{2}{*}{Dialog} & \mrdiffclient & 1.333 & 0.008 & 0.033 & 0.233 & 1.083 \\
             &  & \mrsameclient & 1.146 & 0.000 & 0.033 & 0.179 & 0.946 \\
             \cline{2-8}
             & \multirow{2}{*}{QA} & \mrsameclient  & 0.958 & 0.008 & 0.033 & 0.192 & 0.733 \\
             & & \mrdiffclient  & 0.558 & 0.000 & 0.008 & 0.125 & 0.438 \\
             \cline{2-8}
             & \multirow{2}{*}{Classification} & \mrsameclient & 0.000 & 0.000 & 0.000 & 0.000 & 0.000 \\
             & & \mrdiffclient & 0.000 & 0.000 & 0.000 & 0.000 & 0.000 \\
        \bottomrule
        \end{tabular}
    \end{center}
\end{table*}

We report in Table~\ref{tab:percat_mem_fl_table} \mrsameclient and \mrdiffclient results for each individual category of PAN2014 when used as the discriminator \descf. This provides the breakdown results from our main table (Table~\ref{tab:small_main_table}).
The results show that \mrsameclient is higher than \mrdiffclient, for most categories for Qwen2.5-3B, Llama-3.2-3B, and GPT-2 XL.

\subsection{Trade-off Results between \mrcl and \mrfl}
\label{sec:detail_trade_off_result}
\begin{table*}[h]
\small
\caption{Per-Category Memorization Comparison between CL and FL(\%)}
\label{tab:percat_mem_cl_table}
    \begin{center}
        \begin{tabular}{ccccccc}
        \toprule
             Model & Tasks & Type & Verbatim & Idea & \makecell{Paraphrase\\ (p > 0.5)} & \makecell{Paraphrase\\(p < 0.5)}  \\
             \hline\hline
            \multirow{8}{*}{Qwen2.5-3B} 
             &  \multirow{2}{*}{\makecell{Summarization}} 
             & \mrcl & 0.008 & 0.033 & 0.042 & 0.475 \\ 
             & & \mrfl & 0.008 & 0.000 & 0.050 & 0.375 \\
             \cline{2-7}
             & \multirow{2}{*}{Dialog} 
             & \mrcl & 0.000 & 0.133 & 0.650 & 2.733 \\
             & & \mrfl & 0.000 & 0.142 & 0.817 & 3.217 \\
             \cline{2-7}
             & \multirow{2}{*}{QA} 
             & \mrcl & 0.050 & 0.208 & 0.517 & 1.483 \\
             & & \mrfl & 0.000 & 0.158 & 0.742 & 2.092 \\
             \cline{2-7}
             & \multirow{2}{*}{Classification} 
             & \mrcl & 0.000 & 0.000 & 0.000 & 0.000 \\
             & & \mrfl & 0.000 & 0.000 & 0.000 & 0.000 \\
             \hline
            \multirow{8}{*}{Llama3.2-3B} 
             &  \multirow{2}{*}{\makecell{Summarization}} 
             & \mrcl & 0.017 & 0.025 & 0.108 & 0.792 \\ 
             & & \mrfl & 0.008 & 0.008 & 0.150 & 0.642 \\
             \cline{2-7}
             & \multirow{2}{*}{Dialog} 
             & \mrcl & 0.000 & 0.442 & 0.883 & 4.150 \\
             & & \mrfl & 0.000 & 0.317 & 1.633 & 5.500 \\
             \cline{2-7}
             & \multirow{2}{*}{QA} 
             & \mrcl & 0.092 & 0.267 & 0.783 & 2.525 \\
             & & \mrfl & 0.100 & 0.133 & 1.250 & 3.092 \\
             \cline{2-7}
             & \multirow{2}{*}{Classification} 
             & \mrcl & 0.000 & 0.000 & 0.000 & 0.000 \\
             & & \mrfl & 0.000 & 0.000 & 0.000 & 0.000 \\
        \bottomrule
        \end{tabular}
    \end{center}
\end{table*}
We provide the breakdown results from Table~\ref{tab:small_trade_off} in Table~\ref{tab:percat_mem_cl_table}.

\subsection{Memorization with Various Decoding Methods}
\label{sec:full_decode}
\begin{table*}[ht]
    \vspace{-1ex}
\small
\caption{Memorization with various decoding methods(\%)}
\label{tab:full_decode_table}
    \vspace{-3ex}
    \begin{center}
        \begin{tabular}{lccc|cc|cc|cc} 
        \toprule
\multirow{2}{*}{Model}       & \multirow{2}{*}{\makecell{Decoding}} & \multicolumn{2}{l}{Summarization}                           & \multicolumn{2}{l}{Dialog}                                 & \multicolumn{2}{l}{QA}                          & \multicolumn{2}{l}{Classification}                          \\ \cline{3-10}

                             &                                       & \mrsameclient & \mrdiffclient & \mrsameclient & \mrdiffclient & \mrsameclient & \mrdiffclient & \mrsameclient & \mrdiffclient \\
                             \hline
                             \hline
\multirow{3}{*}{Qwen2.5-3B}  & temperature   & 0.475 & 0.067 & 1.267 & 1.442 & 1.283 & 0.750 & 0.000 & 0.000 \\
                             & top-k         & 0.342 & 0.046 & 1.533 & 1.446 & 1.450 & 0.813 & 0.000 & 0.000 \\
                             & top-p         & 0.525 & 0.050 & 3.792 & 2.996 & 2.567 & 1.438 & 0.000 & 0.000 \\
                             \hline
\multirow{3}{*}{Llama-3.2-3B} & temperature   & 0.442 & 0.117 & 1.400 & 1.096 & 1.458 & 0.925 & 0.000 & 0.000 \\
                             & top-k         & 0.700 & 0.067 & 3.167 & 2.313 & 2.183 & 1.375 & 0.000 & 0.000 \\
                             & top-p         & 0.583 & 0.075 & 3.133 & 2.525 & 2.375 & 1.396 & 0.000 & 0.000 \\
        \bottomrule
        \end{tabular}
    \end{center}
    \vspace{-1ex}
\end{table*}
\mrsameclient and \mrdiffclient tend to increase when top-p or top-k is used in most cases for both Qwen2.5-3B and Llama3.2-3B as shown in Table~\ref{tab:full_decode_table}, suggesting better decoding methods lead to more memorization.

\subsection{Memorization with Various Prefix Lengths}
\label{sec:full_prefix}
\begin{table*}[ht]
\small
\vspace{0ex}

\caption{Memorization with various prefix lengths(\%)}
\label{tab:full_prefix_table}
    \vspace{-3ex}
    \begin{center}
        \begin{tabular}{lccc|cc|cc|cc} 
        \toprule
\multirow{2}{*}{Model}       & \multirow{2}{*}{\makecell{Prefix \\ Length}} & \multicolumn{2}{l}{Summarization}                           & \multicolumn{2}{l}{Dialog}                                 & \multicolumn{2}{l}{Abstractive QA}                          & \multicolumn{2}{l}{Classification}                          \\ \cline{3-10}

                             &                                       & \mrsameclient & \mrdiffclient & \mrsameclient & \mrdiffclient & \mrsameclient & \mrdiffclient & \mrsameclient & \mrdiffclient \\
                             \hline
                             \hline
\multirow{4}{*}{Qwen2.5-3B}  & 10        & 0.508 & 0.188 & 2.108 & 1.992 & 1.525 & 1.383 & 0.000 & 0.000 \\
                             & 30        & 0.342 & 0.046 & 1.533 & 1.446 & 1.450 & 0.813 & 0.000 & 0.000 \\
                             & 50        & 0.425 & 0.038 & 1.575 & 1.242 & 1.242 & 0.550 & - & - \\
                             & 100       & 0.208 & 0.004 & 1.408 & 1.150 & 1.125 & 0.429 & - & - \\
                             \hline
\multirow{4}{*}{Llama-3.2-3B} & 10        & 0.883 & 0.379 & 3.650 & 3.188 & 4.383 & 3.679 & 0.000 & 0.000 \\
                             & 30        & 0.700 & 0.067 & 3.167 & 2.313 & 2.183 & 1.375 & 0.000 & 0.000 \\
                             & 50        & 0.692 & 0.063 & 2.533 & 1.542 & 2.150 & 0.938 & - & - \\
                             & 100       & 0.275 & 0.017 & 3.167 & 2.342 & 2.300 & 0.567 & - & - \\
        \bottomrule
        \end{tabular}
    \end{center}
    \vspace{-2ex}
\end{table*}

Table~\ref{tab:full_prefix_table} shows that increasing prefix length lowers \mrdiffclient and \mrsameclient in most cases for both Qwen2.5-3B and Llama3.2-3B. This phenomenon suggests that shorter prefixes can induce more memorization in FL.

\subsection{Memorization across Federated Algorithms}
\label{sec:full_fl}
\begin{table*}[ht]
\small
\vspace{0ex}

\caption{Memorization with different federated algorithm(\%)}
\label{tab:full_fed_algo_table}
    \vspace{-3ex}
    \begin{center}
        \begin{tabular}{lcccc|c} 
        \toprule
Model  & Tasks     & \makecell{Federated  \\ Algorithm} &  \mrsameclient & \mrdiffclient & Performance \\
                             \hline
                             \hline
\multirow{8}{*}{Qwen2.5-3B}  & \multirow{2}{*}{Summarization}    & FedAvg        &  0.342   & 0.046 & 29.88    \\
                                             & & FedProx     &  0.942   & 0.138 & 30.51    \\
                            \cline{2-6}
                              &  \multirow{2}{*}{Dialog}              & FedAvg        &  1.533   & 1.446 & 18.11   \\
                               &             & FedProx       &  1.892   & 1.879 & 17.47   \\
                            \cline{2-6}
                             &   \multirow{2}{*}{Abstractive QA}             & FedAvg        &  1.450   & 0.813 & 28.60   \\
                              &                  & FedProx       &  3.675   & 2.146 & 28.36   \\
                            \cline{2-6}
                              &   \multirow{2}{*}{Classification}             & FedAvg        &  0.000   & 0.000 & 51.22   \\
                             &                & FedProx       &  0.011   & 0.000 & 80.16   \\
                             \hline
\multirow{8}{*}{Llama-3.2-3B} & \multirow{2}{*}{Summarization}                  & FedAvg        &  0.700   & 0.067 & 28.67     \\
                             &                & FedProx       &  2.442   & 0.413 & 30.78     \\
                            \cline{2-6}
                             &   \multirow{2}{*}{Dialog}             & FedAvg        &  3.167   & 2.313 & 19.44     \\
                            &                 & FedProx       &  8.900   & 6.979 & 18.91     \\
                            \cline{2-6}
                            &     \multirow{2}{*}{Abstractive QA}               & FedAvg        &  2.183   & 1.375 & 28.53     \\
                             &                & FedProx       &  14.092  & 7.892 & 28.76     \\
                            \cline{2-6}
                              &    \multirow{2}{*}{Classification}            & FedAvg        &  0.000   & 0.000 & 78.16     \\
                               &              & FedProx       &  0.300   & 0.117 & 83.14     \\

        \bottomrule
        \end{tabular}
    \end{center}
\vspace{-2ex}
\end{table*}

FedProx has more \mrdiffclient and \mrsameclient than FedAvg as shown in Table~\ref{tab:full_fed_algo_table}.
It also results in memorization in the classification task which is not observed before in FedAvg for both Qwen2.5-3B and Llama3.2-3B. 

\subsection{Memorization with various Model Size}
\label{sec:full_size} 
\begin{table*}[h]
\small
\vspace{0ex}
\caption{Intra- vs inter-client memorization in FL(\%)}
\label{tab:full_size_table}
    \vspace{-3ex}
    \begin{center}
        \begin{tabular}{lcc|cc|cc|cc} 
        \toprule

\multirow{2}{*}{Model} & \multicolumn{2}{l}{Summarization}                           & \multicolumn{2}{l}{Dialog}                                 & \multicolumn{2}{l}{QA}                          & \multicolumn{2}{l}{Classification}                          \\ \cline{2-9}

                       & \mrsameclient & \mrdiffclient & \mrsameclient & \mrdiffclient & \mrsameclient & \mrdiffclient & \mrsameclient & \mrdiffclient \\
                       \hline\hline
Qwen2.5-0.5B & 0.550 & 0.167 & 1.817 & 1.483 & 1.067 & 0.621 & 0.000 & 0.000 \\
Qwen2.5-1.5B & 0.992 & 0.113 & 1.817 & 1.521 & 1.225 & 0.717 & 0.000 & 0.000 \\
Qwen2.5-3B   & 0.342 & 0.046 & 1.533 & 1.446 & 1.450 & 0.813 & 0.000 & 0.000 \\
\hline
Llama-3.2-1B  & 0.800 & 0.088 & 3.808 & 2.938 & 2.108 & 1.275 & 0.000 & 0.000 \\
Llama-3.2-3B  & 0.700 & 0.067 & 3.167 & 2.313 & 2.183 & 1.375 & 0.000 & 0.000 \\ 

        \bottomrule
        \end{tabular}
    \end{center}
\vspace{-2ex}
\end{table*}
As shown in Table~\ref{tab:full_size_table},
we observe no apparent impact of model sizes of both Qwen2.5-3B and Llama3.2-3B on any memorization types across tasks.

\subsection{Memorization with various Communication Rounds}
\label{sec:full_comm_round}
\begin{table*}[ht]
    \vspace{0ex}
\small
\caption{Impact of communication rounds(\%)}
\label{tab:full_com_round_table}
    \vspace{-3ex}
    \begin{center}
        \begin{tabular}{lcccc|c} 
        \toprule
Model  & Tasks     & \makecell{Communication \\ Rounds} &  \mrsameclient & \mrdiffclient & Performance \\
\hline
\hline
\multirow{12}{*}{Qwen2.5-3B}  
    & \multirow{3}{*}{Summarization}    
        & 1 & 0.433 & 0.088 & 29.31 \\
    &   & 3 & 0.342 & 0.046 & 29.88 \\
    &   & 5 & 0.467 & 0.058 & 30.01 \\
\cline{2-6}
    & \multirow{3}{*}{Dialog}    
        & 1 & 1.833 & 1.521 & 18.26 \\
    &   & 3 & 1.533 & 1.446 & 18.11 \\
    &   & 5 & 1.633 & 1.363 & 17.14 \\
\cline{2-6}
    & \multirow{3}{*}{Abstractive QA}    
        & 1 & 1.300 & 0.650 & 28.68 \\
    &   & 3 & 1.450 & 0.813 & 28.60 \\
    &   & 5 & 1.108 & 0.763 & 28.37 \\
\cline{2-6}
    & \multirow{3}{*}{Classification}    
        & 1 & 0.000 & 0.000 & 19.64 \\
    &   & 3 & 0.000 & 0.000 & 51.22 \\
    &   & 5 & 0.000 & 0.000 & 57.57 \\
\hline
\multirow{12}{*}{Llama-3.2-3B}  
    & \multirow{3}{*}{Summarization}    
        & 1 & 0.650 & 0.096 & 27.38 \\
    &   & 3 & 0.700 & 0.067 & 28.67 \\
    &   & 5 & 0.600 & 0.075 & 28.57 \\
\cline{2-6}
    & \multirow{3}{*}{Dialog}    
        & 1 & 4.450 & 3.588 & 19.20 \\
    &   & 3 & 3.167 & 2.313 & 19.44 \\
    &   & 5 & 2.333 & 1.996 & 19.39 \\
\cline{2-6}
    & \multirow{3}{*}{Abstractive QA}    
        & 1 & 2.542 & 1.550 & 28.37 \\
    &   & 3 & 2.183 & 1.375 & 28.53 \\
    &   & 5 & 1.983 & 1.263 & 28.34 \\
\cline{2-6}
    & \multirow{3}{*}{Classification}    
        & 1 & 0.000 & 0.000 & 37.34 \\
    &   & 3 & 0.000 & 0.000 & 78.16 \\
    &   & 5 & 0.000 & 0.000 & 79.73 \\
        \bottomrule
        \end{tabular}
    \end{center}
\vspace{-2ex}
\end{table*}

Table~\ref{tab:full_com_round_table} indicates that communication rounds does not noticeably affect memorization for either Qwen2.5-3B or Llama3.2-3B across tasks.

\subsection{Trade-off between Performance and Memorization CL and FedAvg}
\label{sec:full_trade_off}
\begin{table*}[h]
\small
\vspace{-2ex}
\caption{Trade-off between centralized learning and FedAvg(\%)}
\label{tab:full_trade_off}
\vspace{-3ex}
    \begin{center}
        \begin{tabular}{lcccc|c} 
        \toprule
                       Model & Task &  Framework  & Bleu & RougeL & Memorization  \\
                       \hline
                       \hline
\multirow{9}{*}{Qwen2.5-3B} & \multirow{2}{*}{ \makecell{Summarization}}   & \mrcl & 37.04 & 28.46 & 0.558 \\
                       & & \mrfl  & 37.20 & 29.88  & 0.433 \\

                       \cline{2-6}
& \multirow{2}{*}{Dialog}   & \mrcl & 30.55 & 19.40 & 3.417 \\
                       & & \mrfl  & 28.87 & 18.11 & 3.992 \\

                       \cline{2-6}
& \multirow{2}{*}{QA}   & \mrcl & 32.47 & 26.66 & 2.150 \\
                       & & \mrfl & 31.25 & 28.60 & 2.917 \\
                        \cline{2-6}
                        & &    & Accuracy &  &  \\
                       \cline{2-6}
& \multirow{2}{*}{Classification}   & \mrcl & 76.30 &  & 0.000 \\
                       & & \mrfl & 51.22 & &  0.000 \\
                       \hline

\multirow{9}{*}{Llama-3.2-3B} & \multirow{2}{*}{ \makecell{Summarization}}   & \mrcl & 35.03 & 26.80 & 0.942 \\
                       & & \mrfl  & 36.95 & 28.67 & 0.808 \\

                       \cline{2-6}
& \multirow{2}{*}{Dialog}   & \mrcl & 31.05 & 19.62 & 5.308 \\
                       & & \mrfl  & 29.77 & 19.44 & 7.050 \\

                       \cline{2-6}
& \multirow{2}{*}{QA}   & \mrcl & 32.80 & 26.67  & 3.458  \\
                       & & \mrfl & 32.94 & 28.53 & 4.300 \\
                        \cline{2-6}
                        & &    & Accuracy &  &  \\
                       \cline{2-6}
& \multirow{2}{*}{Classification}   & \mrcl & 77.83 &  & 0.000 \\
                       & & \mrfl & 78.16 &  &  0.000  \\
        \bottomrule
        \end{tabular}
    \end{center}
\vspace{-2ex}
\end{table*}

We observe no clear trend in memorization when shifting from CL to FL for Qwen2.5-3B or Llama3.2-3B as shown in Table~\ref{tab:full_trade_off}. Furthermore, we found no correlation between memorization and performance in either setting.

\end{document}